\documentclass[11pt,a4paper]{article}
\usepackage[hyperref]{acl2020}
\usepackage{times}
\usepackage{latexsym}

\newcommand\blfootnote[1]{%
\begingroup
\renewcommand\thefootnote{}\footnote{#1}%
\addtocounter{footnote}{-1}%
\endgroup
}

\usepackage{microtype}
\usepackage{enumitem}
\usepackage{color}
\usepackage{amsmath}
\usepackage{algorithm}
\usepackage{algorithmic}
\usepackage{graphicx}
\usepackage{lipsum}
\aclfinalcopy % Uncomment this line for the final submission
\def\aclpaperid{688} %  Enter the acl Paper ID here

\title{A Batch Normalized Inference Network Keeps the KL Vanishing Away}

\author{Qile Zhu\textsuperscript{1}$^*$ , Jianlin Su$^*$, Wei Bi\textsuperscript{2}, Xiaojiang Liu\textsuperscript{2}, Xiyao Ma\textsuperscript{1}, Xiaolin Li\textsuperscript{3} and Dapeng Wu\textsuperscript{1}
 \\
   \textsuperscript{1}University of Florida, \textsuperscript{2}Tencent AI Lab, \textsuperscript{3}AI Institute, Tongdun Technology \\
  \tt {\{valder,maxiy,dpwu\}@ufl.edu} \\ \tt {\{victoriabi,kieranliu\}@tencent.com}\\ \tt  xiaolin.li@tongdun.net \\
  \tt bojone@spaces.ac.cn}

\date{}

\begin{document}
\maketitle
\begin{abstract}
\blfootnote{*This work was done when Qile Zhu was an intern at Tencent AI Lab. Wei Bi is the corresponding author. Jianlin Su further extends this work.}Variational Autoencoder (VAE) is widely used as a generative model to approximate a model's posterior on latent variables by combining the amortized variational inference and deep neural networks. However, when paired with strong autoregressive decoders, VAE often converges to a degenerated local optimum known as ``posterior collapse''. Previous approaches consider the Kullback–Leibler divergence (KL) individual for each datapoint. We propose to let the KL follow a distribution across the whole dataset, and analyze that it is sufficient to prevent posterior collapse by keeping the expectation of the KL's distribution positive. Then we propose Batch Normalized-VAE (BN-VAE), a simple but effective approach to set a lower bound of the expectation by regularizing the distribution of the approximate posterior's parameters. Without introducing any new model component or modifying the objective, our approach can avoid the posterior collapse effectively and efficiently. We further show that the proposed BN-VAE can be extended to conditional VAE (CVAE). Empirically, our approach surpasses strong autoregressive baselines on language modeling, text classification and dialogue generation, and rivals more complex approaches while keeping almost the same training time as VAE. 
\end{abstract}

\section{Introduction}
Variational Autoencoder (VAE) \cite{kingma2013auto,rezende2014stochastic}is one of the most popular generative framework to model complex distributions. Different from the Autoencoder (AE), VAE provides a distribution-based latent representation for the data, which encodes the input $\mathbf{x}$ into a probability distribution $\mathbf{z}$ and reconstructs the original input using samples from $\mathbf{z}$. When inference, VAE first samples the latent variable from the prior distribution and then feeds it into the decoder to generate an instance. VAE has been successfully applied in many NLP tasks, including topic modeling \cite{srivastava2017autoencoding,miao2016neural,zhu2018graphbtm}, language modeling \cite{bowman2015generating}, text generation \cite{zhao2017learning} and text classification \cite{xu2017variational}. 

An autoregressive decoder (e.g., a recurrent neural network) is a common choice to model the text data. However, when paired with strong autoregressive decoders such as LSTMs \cite{hochreiter1997long} and trained under conventional training strategy, VAE suffers from a well-known problem named the \textit{posterior collapse} or the KL \textit{vanishing} problem. The decoder in VAE learns to reconstruct the data independent of the latent variable $\mathbf{z}$, and the KL vanishes to 0.

    Many convincing solutions have been proposed to prevent posterior collapse. Among them, fixing the KL as a positive constant is an important direction~\cite{davidson2018hyperspherical,guu2018generating,oord2017neural,xu2018spherical,tomczak2017vae,kingma2016improved,razavi2018preventing}. Some change the Gaussian prior with other distributions, e.g., a uniform prior \cite{oord2017neural, zhao2018unsupervised} or a von Mises-Fisher (vMf) distribution \cite{davidson2018hyperspherical,guu2018generating,xu2018spherical}. However, these approaches force the same constant KL and lose the flexibility to allow various KLs for different data points \cite{razavi2018preventing}. Without changing the Gaussian prior, free-bits \cite{kingma2016improved} adds a threshold (free-bits) of the KL term in the ELBO object and stops the optimization of the KL part when its value is smaller than the threshold. \citet{chen2016variational} point out that the objective of free-bits is non-smooth and suffers from the optimization challenges. $\delta$-VAE \cite{razavi2018preventing} sets the parameters in a specific range to achieve a positive KL value for every latent dimension, which may limit the model performance.
    
    Other work analyzes this problem form a view of optimization \cite{bowman2015generating,zhao2017infovae,chen2016variational, alemi2017fixing}. Recently, \citet{he2019lagging} observe that the inference network is lagging far behind the decoder during training. They propose to add additional training loops for the inference network only. \citet{li-etal-2019-surprisingly} further propose to initialize the inference network with an encoder pretrained from an AE objective, then trains the VAE with the free-bits. However, these two methods are much slower than the original VAE. 
    
    The limitation of the constant KL and the high cost of additional training motivate us to seek an approach that allows flexible modeling for different data points while keeping as fast as the original VAE. In this paper, instead of considering the KL individually for each data point, we let it follow a distribution across the whole dataset. We demonstrate that keeping a positive expectation of the KL's distribution is sufficient to prevent posterior collapse in practice. By regularizing the distribution of the approximate posterior's parameters, a positive lower bound of this expectation could be ensured. Then we propose Batch Normalized-VAE (BN-VAE), a simple yet effective approach to achieving this goal, and discuss the connections between BN-VAE and previous enhanced VAE variants.     We further extend BN-VAE to the conditional VAE (CVAE). Last, experimental results demonstrate the effectiveness of our approach on real applications, including language modeling, text classification and dialogue generation. Empirically, our approach surpasses strong autoregressive baselines and is competitive with more sophisticated approaches while keeping extremely higher efficiency. Code and data are available at \url{https://github.com/valdersoul/bn-vae}.

\section{Background and Related Work}
    \label{sec:background}
    %\subsection{VARIATIONAL AUTOENCODERS}
    
    In this section, we first introduce the basic background of VAE, then we discuss the lagging problem \cite{he2019lagging}. At last, we present more related work.
    
    \subsection{VAE Background} 
    VAE \cite{kingma2013auto,rezende2014stochastic} aims to learn a generative model $p(\mathbf{x},\mathbf{z})$ to maximize the marginal likelihood $\log p(\mathbf{x})$ on a dataset. The marginal likelihood cannot be calculated directly due to an intractable integral over the latent variable $\mathbf{z}$. To solve this, VAE introduces a variational distribution $q_{\phi}(\mathbf{z}|\mathbf{x})$ which is parameterized by a complex neural network to approximate the true posterior. Then it turns out to optimize the ELBO of $\log p(\mathbf{x})$:
    \begin{align}
    \mathcal{L} = \mbox{E}_{q_{\phi}(\mathbf{z}|\mathbf{x})}[\log p_{\theta}(\mathbf{x}|\mathbf{z})] - KL(q_{\phi}(\mathbf{z}|\mathbf{x})||p(\mathbf{z})),
    \end{align}
    where $\mathbf{\phi}$ represents the inference network and $\mathbf{\theta}$ denotes the decoder. The above first term is the reconstruction loss, while the second one is the KL between the approximate posterior and the prior. The Gaussian distribution $\mathcal{N} \sim (0, \textit{I})$ is a usual choice for the prior, and the KL between the approximate posterior $q_{\phi}(\mathbf{z}|\mathbf{x})$ and the prior $p(\mathbf{z})$ can be computed as:
    \begin{align}\label{eq:3}
    KL = \frac{1}{2}  \sum_{i=1}^n(\mu_i^2 + \sigma_i^2 - \log\sigma_i^2 - 1),
    \end{align}
    where $\mu_i$ and $\sigma_i$ is the mean and standard deviation of approximate posterior for the $i_{th}$ latent dimension, respectively. When the decoder is autoregressive, it can recover the data independent of the latent $\mathbf{z}$ \cite{bowman2015generating}. The optimization will encourage the approximate posterior to approach the prior which results to the zero value of the KL. 
    \subsection{The Lagging Problem} 
    \label{sec:lagging_related}
    Recently, \citeauthor{he2019lagging}~\shortcite{he2019lagging} analyze posterior collapse with the Gaussian prior from a view of training dynamics. The collapse is a local optimum of VAE when $q_{\phi}(\mathbf{z}|\mathbf{x}) = p_{\theta}(\mathbf{z}|\mathbf{x})=p(\mathbf{z})$ for all inputs. They further define two partial collapse states: \textit{model collapse}, when $p_{\theta}(\mathbf{z}|\mathbf{x})=p(\mathbf{z})$, and \textit{inference collapse}, when $q_{\phi}(\mathbf{z}|\mathbf{x})=p(\mathbf{z})$. They observe that the inference collapse always happens far before the model collapse
    % inference network $q_\phi(\mathbf{z}|\mathbf{x})$ is lagging far behind the model posterior $p_\theta(\mathbf{z}|\mathbf{x})$ 
    due to the existence of autoregressive decoders. Different from the model posterior, the inference network lacks of guidance and easily collapses to the prior at the initial stage of training, and thus posterior collapse happens. Based on this understanding, they propose to aggressively optimize the inference network. However, this approach cost too much time compared with the original VAE.
    In our work, we also employ the Gaussian prior and thus suffer from the same lagging problem. Yet, our proposed approach does not involve additional training efforts, which can effectively avoid the lagging problem (Section~\ref{sec:lagging}) and keep almost the same training efficiency as the original VAE (Section~\ref{sec:lm}). 
    More details can be found in Section~\ref{sec:lagging}. 

    \subsection{Related Work} 
    % \footnote{***where is the lagging paper?}
    \label{sec:related}
    To prevent posterior collapse, we have mentioned many work about changing the prior in the introduction. Besides these approaches, some work modifies the original training objective directly. %\cite{alemi2017fixing,bowman2015generating,higgins2017beta,tschannen2018recent,zhao2017infovae} 
    %or weakens the decoder \cite{bowman2015generating,semeniuta2017hybrid,yang2017improved} . 
    %\footnote{***bowman2015 is cited twice in this sentence.}
    For example, \citet{bowman2015generating} introduce an annealing strategy, where they slightly increase the weight of KL from 0 to 1 during the warm-up period. $\beta$-VAE~\cite{higgins2017beta} treats the KL weight as a hyperparameter to constrain the minimum value of the KL. 
    \citet{alemi2016deep}, on the other hand, set a fixed KL weight to control the mutual information between $\mathbf{z}$ and $\mathbf{x}$. \citet{tolstikhin2017wasserstein} leverage the wasserstein distance to replace the KL. \citet{zhao2017infovae} replace the KL with maximum mean discrepancy. \citet{fang2019implicit} introduce sample-based representations which lead to implicit latent features with an auxiliary network.
    
    Some change the training strategy. \citet{kim2018semi} address the amortization gap \cite{pmlr-v80-cremer18a} in VAE and propose Semi-Amortized VAE to compose the inference network with additional mean-field updates.
    \citet{liu2019cyclical} propose a cyclical annealing schedule, which repeats the process of increasing $\beta$ multiple times.
    
    There are various other approaches to solve the posterior collapse. For example, some researchers choose to weaken the decoder
  %  \citeauthor{semeniuta2017hybrid}~\shortcite{semeniuta2017hybrid} and \citeauthor{yang2017improved}~\shortcite{yang2017improved}
  by replacing the LSTM decoder with convolution neural networks without autoregressive modeling~\cite{semeniuta2017hybrid,yang2017improved}.
  \citet{chen2016variational} input a lossy representation of data to the autoregressive decoder  and enforce $\mathbf{z}$ to capture the information about the original input. Inheriting this idea, some following work add direct connections between $\mathbf{z}$ and $\mathbf{x}$~\cite{zhao2017learning,dieng2018avoiding}. \citet{ma2019mae} introduce an additional regularization to learn diverse latent representation. $\delta$-VAE \cite{razavi2018preventing} and free-bits \cite{kingma2016improved} set a minimum number of KL for each latent dimension to prevent the posterior collapse.

    %Although BN has been applied in VAE in several previous work, they aim to solve totally different problems.
    \citet{srivastava2017autoencoding,srivastava2018variational} find that using ADAM \cite{kingma2014adam} with a high learning rate to train VAE may cause the gradients to diverge early. Their explanation for the diverging behavior lies in the exponential curvature of the gradient from the inference network which produces the variance part of the approximate posterior. Then they apply batch normalization to the variance part to solve this problem. We use the simple SGD without momentum to train our model. Moreover, we apply batch normalization to the mean part of the inference network to keep the expectation of the KL's distribution positive, which is different from their work. We also find that \citet{sonderby2016ladder} utilize batch normalization in all fully connected layers with nonlinear activation functions to improve the model performance. Different from it, our approach directly applies batch normalization to the parameters of the approximate posterior, which is the output of the inference network. 
    
    % \begin{figure*}
    %     \centering
    %     \includegraphics[width=0.8\textwidth]{figs/synthetic_v4.pdf}
    %     \caption{Visualization of 500 sampled data from the synthetic dataset during the training. The x-axis is $\mu_{x,\theta}$, the approximate model posterior mean. The y-axis is $\mu_{x,\phi}$, which represents the inference posterior mean. b is batch size and $\gamma$ is 1 in BN.}
    %     \label{fig:synthetic}
    % \end{figure*}
    
    \section{Batch-Normalized VAE}
    In this section, we first derive the expectation of the KL's distribution and show that it is enough to avoid posterior collapse by keeping the expectation of the KL's distribution positive. Then we propose our regularization method on the parameters of the approximate posterior to ensure a positive lower bound of this expectation. 
    We further discuss the difference between our approach and previous work.
    % , including free-bits \cite{kingma2016improved}, $\delta$-VAE \cite{razavi2018preventing} and our approach can overcome the lagging problem. 
    %At last, we extend our approach to CVAE.
    % In this section, we introduce our approach to overcome posterior collapse. We first present how we derive the expectation of the KL's distribution ($D_{KL}$) and explain why we only need to care about it. Then we propose our regularization method on the parameters of the approximate posterior to ensure a positive lower bound of the expectation of the $D_{KL}$ with the Gaussian prior. We further show the connections between our approach and previous ones. At last, we show the extension of our approach to CVAE.
    %Then an aggressive training strategy is proposed to solve this problem \cite{he2019lagging}, but it needs much more time compared with the original VAE training.
    
    \subsection{Expectation of the KL's Distribution}
    \label{sec:mean}
    
    Given an $\boldsymbol{x} \in \mathcal{X}$, the inference network parametrizes a $n$-dimension diagonal Gaussian distribution with its mean $\mathbf{\mu}=f_{\mu}(\boldsymbol{x})$ and diagonal covariance $\mathbf{\Sigma}=diag(f_{\Sigma}(\boldsymbol{x}))$, where $f_{\mu}$ and $f_{\Sigma}$ are two neural networks. In practice, the ELBO is computed through a Monte Carlo estimation from $b$ samples. The KL in Eq.~\ref{eq:3} is then computed over $b$ samples from $\mathcal{X}$:
    \begin{align}
    \label{eq:KL_dis}  
    KL  & =   \frac{1}{2b}  \sum_{j=1}^b\sum_{i=1}^n(\mu_{ij}^2 + \sigma_{ij}^2 - \log\sigma_{ij}^2 - 1) \nonumber\\
    & =  \frac{1}{2}\sum_{i=1}^n(\frac{\sum_{j=1}^b\mu_{ij}^2}{b} + \frac{\sum_{j=1}^b\sigma_{ij}^2}{b} \nonumber \\ 
    & -  \frac{\sum_{j=1}^b\log\sigma_{ij}^2}{b} - 1). 
    \end{align} 
    %where $n$ denotes the number of latent dimensions.
    When $b$ gets larger, the above empirical value will approach the mean of the KL across the whole dataset. 
    
    To make use of this observation, we assume that $\mu_i$ and $\log\sigma_i^2$ for each latent dimension $i$ follow a certain distribution with a fixed mean and variance across the dataset respectively. The distribution may vary between different latent dimensions. In this way, the KL turns to a distribution of $\mu_i$'s and $\log\sigma_i^2$'s. From Eq.~\ref{eq:KL_dis}, we can see that $\sum_{j=1}^b\mu_{ij}^2 / b$ is the sample mean of $\mu_i^2$, which converges to $\mbox{E}[\mu_i^2] = \mbox{Var}[\mu_i]+\mbox{E}^2[\mu_i]$. 
    Similarly, $\sum_{j=1}^b\sigma_{ij}^2 / b$ converges to $\mbox{E}[\sigma_{i}^2]$, and $\sum_{j=1}^b \log \sigma_{ij}^2/b$ to $\mbox{E}[\log\sigma_i^2]$. 
    Thus, we can derive the expectation of the KL's distribution as:
    \begin{align}
    \mbox{E}[KL] & = \frac{1}{2}\sum_{i=1}^n(\mbox{Var}[\mu_i]+\mbox{E}^2[\mu_i]  \nonumber \\ 
    & + \mbox{E}[\sigma_{i}^2] - \mbox{E}[\log\sigma_i^2] - 1) \nonumber\\
    & \geq \frac{1}{2}\sum_{i=1}^n(\mbox{Var}[\mu_i] + \mbox{E}^2[\mu_i]),\label{eq:kl7}
    \end{align}
    where $\mbox{E}[\sigma_i^2 - \log\sigma_i^2] \geq 1$ since the minimum of $e^x -x$ is 1. 
    If we can guarantee a positive lower bound of $\mbox{E}[KL]$, we can then effectively prevent the posterior collapse.
    
    Based on Eq.~\ref{eq:kl7}, the lower bound is only dependent on the number of latent dimensions $n$ and $\mu_i$'s mean and variance. This motivates our idea that with proper regularization on the distributions of $\mu_i$'s to ensure a positive lower bound of $\mbox{E}[KL]$.
    
    \subsection{Normalizing Parameters of the Posterior}
    \label{sec:bn}
    The remaining key problem is to construct proper distributions of $\mu_i$'s that can result in a positive lower bound of $\mbox{E}[KL]$ in Eq.~\ref{eq:kl7}. 
     %From Eq.~\ref{eq:kl7},  we only need to care about $\mu_i$'s mean and variance in the dataset. 
     %Thus, we propose to apply BN on the output of the inference network ($\mu_i$).
    Here, we propose a simple and efficient approach to accomplish this by applying a fixed batch normalization on the output of the inference network ($\mu_i$).
    Batch Normalization (BN) \cite{ioffe2015batch} is a widely used regularization technique in deep learning. It normalizes the output of neurons and makes the optimization landscape significantly smoother~\cite{santurkar2018does}. Different from other tasks that apply BN in the hidden layers and seek fast and stable training, here we leverage BN as a tool to transform $\mu_i$ into a distribution with a fixed mean and variance. Mathematically, the regularized $\mu_i$ is written by:
    \begin{align} \label{eq:7}
    \hat{\mu_i}=\gamma\frac{\mu_i-\mu_{\mathcal{B}i}}{\sigma_{\mathcal{B}i}} + \beta,
    \end{align}
    where $\mu_i$ and $\hat{\mu_i}$ are means of the approximate posterior before and after BN. $\mu_{\mathcal{B}i}$ and $\sigma_{\mathcal{B}i}$ denote the mean and standard deviations of $\mu_i$. They are biased estimated within a batch of samples for each dimension indecently. $\gamma$ and $\beta$ are the scale and shift parameter. Instead of using a learnable $\gamma$ in Eq. \ref{eq:7}, we use a fixed BN which freezes the scale $\gamma$. In this way, the distribution of $\mu_i$ has the mean of $\beta$ and the variance of $\gamma^{2}$. $\beta$ is a learnable parameter that makes the distribution more flexible. 
    
    Now, we derive the lower bound of $\mbox{E}[KL]$ by using the fixed BN. With the fixed mean $\beta$ and variance $\gamma^{2}$ for $\mu_i$ in hand, we get a new lower bound as below:
    \begin{align}\label{eq:gamma}
    \mbox{E}[KL]  & \geq \frac{1}{2}\sum_i^n(\mbox{Var}[\mu_i] + \mbox{E}^2[\mu_i]) \nonumber \\ 
    & = \frac{n \cdot (\gamma^{2}+\beta^{2})}{2}. 
    \end{align}
    To this end, we can easily control the lower bound of $\mbox{E}[KL]$ by setting $\gamma$. Algorithm \ref{alg:bnvae} shows the training process.
    %By applying this fixed BN, we also incorporate the Gaussian as the prior of $q_{\phi}(z|x)$. The lagging problem is mainly due to the lack of guidance for the inference network at the initial training stage. This prior is a strong signal to force the inference network far wary from the prior (see Sec \ref{sec:synthetic}).
    \begin{algorithm}
    \caption{BN-VAE training.}
    \renewcommand{\algorithmicrequire}{ \textbf{Input:}} %Use Input in the format of Algorithm  
    \renewcommand{\algorithmicensure}{ \textbf{Output:}} %Use Output in the format of Algorithm  
    \label{alg:bnvae}  
    \begin{algorithmic}[1]    
    % \REQUIRE ~~\\              
    %   Training set $\mathcal{D}$ %= \{ x^{1}, x^{2},\dotsb, x^{N} \}$
    % \ENSURE ~~\\                 
    %   Encoder $\boldsymbol{\phi}$ and autoregressive decoder $\boldsymbol{\theta}$
    
    \STATE Initialize $\boldsymbol{\phi}$ and $\boldsymbol{\theta}$.
    \FOR {$i=1,2,\dotsb$ Until Convergence}
        \STATE Sample a mini-batch $\mathbf{x}$.
        \STATE $\mathbf{\mu}$, $\log \mathbf{\sigma}^2$ = $f_{\phi}(\mathbf{x})$.
        \STATE $\mathbf{\mu'}$ = $BN_{\gamma, \beta}(\mathbf{\mu})$.
        \STATE Sample $\mathbf{z} \sim \mathcal{N}(\mathbf{\mu', \sigma}^2)$ and reconstruct $\mathbf{x}$ from $f_\theta(\mathbf{z})$.
        \STATE Compute gradients $\mathbf{g}_{\phi,\theta} \leftarrow{} \nabla_{\phi,\theta}\mathcal{L}(\mathbf{x};\phi,\theta)$.
        \STATE Update $\phi, \theta$ using $\mathbf{g}_{\phi,\theta}$.
    \ENDFOR
    \end{algorithmic}
    \end{algorithm}
    
    % \subsection{Discuss with free-bits and $\delta$-VAE}

\subsection{Further Extension \footnote{Thanks Jianlin Su for this part! More details can be chekc on \url{https://kexue.fm/archives/7381} (in Chinese, but code is the universal language!)}}
In this section, we will derive a more strict form for our algorithm. The ultimate goal of VAE is to mapping the posterior to the prior if possible. Then we have:
\begin{align} 
    q(z)= & \int \hat{p}(x)p(z|x)dx  \\
    = & \int \hat{p}(x)\mathcal{N}(z;\mu(x),\sigma(x))dx.
\end{align}
If we multiply $z$ and integrate with $z$ on both sides, we can get:
\begin{align}
    0=\int \hat{p}(x)\mu(x)dx = \mbox{E}_{x \sim \hat{p}(x)}[\mu(x)].
\end{align}
If we multiply $z^2$ and do the same thing, we can get:
\begin{align}
    1= & \int \hat{p}(x)[\mu(x)^2 + \sigma(x)^2]dx \\
    = & \mbox{E}_{x \sim \hat{p}(x)}[\mu(x)^2] + \mbox{E}_{x \sim \hat{p}(x)}[\sigma(x)^2]
\end{align}
By this end, we can apply BN to both $\mu$ and $\sigma$ parts, we have:
\begin{align}
    \beta_{\mu} = 0 
\end{align}
and:
\begin{align}
        \beta_{\mu}^2 + \gamma_{\mu}^2 + \beta_{\sigma}^2 + \gamma_{\sigma}^2 = 1
\end{align}
We can further fix $\beta_{\sigma}=0$. To this end, we propose another BN-VAE based on this part:
\begin{align}
    & \beta_{\mu} = \beta_{\sigma} = 0 \\
    & \gamma_{\mu} = \sqrt{\tau+(1-\tau)\cdot sigmoid(\theta)} \\
    & \gamma_{\sigma} = \sqrt{(1-\tau)\cdot sigmoid(-\theta)},
\end{align}
where $\tau \in (0,1)$ and $\theta$ is a trainable parameter.
\subsection{Connections with Previous Approaches}
\label{sec:lagging}
    {\bf Constructing a positive KL:} Both free-bits \cite{kingma2016improved} and $\delta$-VAE \cite{razavi2018preventing} set a threshold on the KL value. Free-bits changes the KL term in the ELBO to a hinge loss term:
    %\begin{align}
        $\sum_{i}^n \max(\lambda, KL(q_{\phi}(z_i|x)||p(z_i)))$.
    %\end{align}
    Another version of free-bits is to apply the threshold to the entire sum directly instead of the individual value. Training with the free-bits objective, the model will stop to drive down the KL value when it is already below $\lambda$. However, \citet{chen2016variational} point out that the objective of free-bits is non-smooth and suffers from the optimization challenges. Our approach does not face the optimization problem since we use the original ELBO objective. 
    
    $\delta$-VAE sets a target rate of $\delta$ for each latent dimension by constraining the mean and variance of the approximate posterior:
    \begin{align}
        \sigma_q=\sigma_q^l+(\sigma_q^u - \sigma_q^l)\frac{1}{1+e^{-q_{\phi}(x)}},
    \end{align}
    \begin{align}
        \mu=2\delta+1+\ln(\sigma_q^2)-\sigma_q^2+\max(0,\mu_{\phi}(\mathbf{x})),
    \end{align}
    where [$\sigma^l$, $\sigma^u$] are the feasible interval for $\sigma_q$ by solving $\ln(\sigma_q^2) - \sigma_q^2 + 2\delta+1 \geq 0$. Although $\delta$-VAE can ensure a minimum value for the KL, it limits the model performance due to that the parameters are constrained in the interval. Our approach only constrains the distributions of $\mu$, which is more flexible than $\delta$-VAE. Experiments further show that our approach surpass both free-bits and $\delta$-VAE. 

    \noindent
    {\bf Reducing inference lag:}
    As we focus on the setting of the conventional Gaussian prior, the lagging problem mentioned in Section~\ref{sec:lagging_related} is crucial. To this point, it is beneficial to analyze an alternate form of the ELBO:
    \begin{align} \label{eq9}
    \mathcal{L} =  \log p_{\theta}(\mathbf{x}) - KL(q_{\phi}(\mathbf{z}|\mathbf{x})||p_{\theta}(\mathbf{z}|\mathbf{x})).
    \end{align}
    With this view, the only goal of the approximate posterior $q_{\phi}(\mathbf{z}|\mathbf{x})$ is to match the model posterior $p_{\theta}(\mathbf{z}|\mathbf{x})$. 
    % As presented in Sec \ref{sec:background}, the lagging problem is correlated to posterior collapse. To test our approach, we generate discrete synthetic data since posterior collapse usually happens in text modeling tasks \cite{he2019lagging}. 
    % The VAE used in this toy task has a LSTM encoder and a LSTM decoder. We use a scalar latent variable because we need to compute $\mu_{x,\theta}$ approximated by discretization of $p_{\theta}(z|x)$. To visualize the training progress, we sample 500 data points from validation set and show them on the mean space of inference and model posteriors as in Fig. \ref{fig:synthetic}.
    We examine the performance of our approach to reduce inference lag using the same synthetic experiment in~\citet{he2019lagging}. Details can be found in Section 1 of the Appendix.
    % We plot the mean value of the approximate posterior and the model posterior during training for the basic VAE and BN-VAE. As shown the first column in Fig.~\ref{fig:synthetic}, all points have the zero mean of the model posterior (the x-axis), which indicates that $\mathbf{z}$ and $\mathbf{x}$ are independent at the beginning of training. For the basic VAE, points start to spread in the x-axis during training while sharing almost the same y value, since the model posterior $p_\theta(\mathbf{z}|\mathbf{x})$ is well learned with the help of the autoregressive decoder. However, the inference posterior $q_\phi(\mathbf{z}|\mathbf{x})$ is lagging behind $p_\theta(\mathbf{z}|\mathbf{x})$ and collapses to the prior in the end. Our regularization approximated by BN, on the other hand, pushes the inference posterior $q_{\phi}(\mathbf{z}|\mathbf{x})$ away from the prior ($p(\mathbf{z})$) at the initial training stage, and forces $q_\phi(\mathbf{z}|\mathbf{x})$ to catch up with $p_\theta(\mathbf{z}|\mathbf{x})$ to minimize $KL(q_{\phi}(\mathbf{z}|\mathbf{x})||p_{\theta}(\mathbf{z}|\mathbf{x}))$ in Eq.~\ref{eq9}. As in the second row of Fig.~\ref{fig:synthetic}, points spread in both directions and towards the diagonal in the end. 
    The synthetic experiment indicates that our approach with the regularization is beneficial to rebalance the optimization between inference and generation, and finally overcomes posterior collapse. We also prefer a large $\gamma$ due to that a small $\gamma$ will push the approximate posterior to the prior. More details on the synthetic experiment can be found in the Appendix.
    
    \section{Extension to CVAE}
    Given an observation $\boldsymbol{\mathbf{x}}$ and its output $\boldsymbol{\mathbf{y}}$, CVAE \cite{NIPS2015_5775,zhao2017learning} models the conditional distribution $p(\mathbf{y}|\mathbf{x})$. The variational lower bound of the conditional log-likelihood is:
    \begin{align}
    % \mathcal{L}(\mathbf{x},\mathbf{y};\theta,\phi) =
    \mathcal{L} & = \mbox{E}_{q_{\phi}(\mathbf{z}|\mathbf{x},\mathbf{y})}[\log p_{\kappa}(\mathbf{y}|\mathbf{x},\mathbf{z})] \label{eq:cdlbo} \nonumber \\
    & -KL(q_{\phi}(\mathbf{z}|\mathbf{x},\mathbf{y})||p_{\theta}(\mathbf{z}|\mathbf{x})) \nonumber\\
    & \leq \log p(\mathbf{y}|\mathbf{x}).
    \end{align}
    Different from VAE, the prior $p_\theta(\mathbf{z}|\mathbf{x})$ in CVAE is not fixed, which is also parametrized by a neural network. It is possible to apply another BN on the mean of the prior with a different $\gamma$ so that the expectation of the KL becomes a constant. However, 
    this lower bound is uncontrollable due to the density of $\mu_1 + \mu_2$ is the convolution of their densities, which is intractable. \footnote{We perform empirical study on this method and find that the neural network can always find a small $KL$ value in this situation.}

    To overcome this issue, we propose to constrain the prior with a fixed distribution. We achieve it by adding another KL between the prior and a known Gaussian distribution $r(\mathbf{z})$, i.e. $KL(p_{\theta}(\mathbf{z}|\mathbf{x}) || r(\mathbf{z}))$. Instead of optimizing the ELBO in Eq.~\ref{eq:cdlbo}, we optimize a lower bound of the ELBO for CVAE:
    \begin{align}
    \mathcal{L}' = \mathcal{L} - KL(p_{\theta}(\mathbf{z}|\mathbf{x}) || r(\mathbf{z})) 
     \leq \mathcal{L}.
    \end{align}
    % In this way, we can apply BN on the inference network $q_\phi(\mathbf{z}|\mathbf{x},\mathbf{y})$ and avoid posterior collapse. \footnote{***dont understand this, also try to give an algo.}
    The KL term in the new bound is the sum of $KL(q_{\phi}(\mathbf{z}|\mathbf{x},\mathbf{y})||p_{\theta}(\mathbf{z}|\mathbf{x}))$ and $KL(p_{\theta}(z|x)||r(z))$, which can be computed as:
    \begin{align}
        KL&=\frac{1}{2}\sum_{i=1}^n(\frac{\sigma_{qi}^2 + (\mu_{qi}-\mu_{pi})^2}{\sigma_{pi}^2} \nonumber \\
        & +\sigma_{pi}^2+\mu_{pi}^2-log\sigma_{qi}^2-1),
    \end{align}
    where $\sigma_{q}$, $\mu_{q}$ and $\sigma_{p}$, $\mu_{p}$ are the parameters of $q_{\phi}$ and $p_{\theta}$ respectively. $n$ denotes the hidden size. The KL term vanishes to 0 when and only when $q_{\phi}$ and $p_{\theta}$ collapse to $r(\mathbf{z})$, which is the normal distribution. As we explained in Section~\ref{sec:bn}, KL won't be 0 when we apply BN in $q_{\phi}$. We then prove that when $q_{\phi}$ collapses to $p_{\theta}$, the KL term is not the minima (details in Section 2 of the Appendix) so that $KL(q_{\phi}(\mathbf{z}|\mathbf{x},\mathbf{y})||p_{\theta}(\mathbf{z}|\mathbf{x}))$ won't be 0. In this way, we can avoid the posterior collapse in CVAE. Algorithm \ref{alg:bncvae} shows the training details.
    
    \begin{algorithm}
    \caption{BN-CVAE training.}
    \renewcommand{\algorithmicrequire}{ \textbf{Input:}} %Use Input in the format of Algorithm  
    \renewcommand{\algorithmicensure}{ \textbf{Output:}} %Use Output in the format of Algorithm  
    \label{alg:bncvae}  
    \begin{algorithmic}[1]    
    % \REQUIRE ~~\\              
    %   Training set $\mathcal{D}$ %= \{ x^{1}, x^{2},\dotsb, x^{N} \}$
    % \ENSURE ~~\\                 
    %   Encoder $\boldsymbol{\phi}$, prior network $\boldsymbol{\theta}$and decoder $\boldsymbol{\kappa}$
    
    \STATE Initialize $\phi$, $\theta$ and $\kappa$.
    \FOR {$i=1,2,\dotsb$ Until Convergence}
        \STATE Sample a mini-batch $\mathbf{x},\mathbf{y}$.
        \STATE $\mathbf{\mu}_q$, $\log \mathbf{\sigma}_q^2$ = $f_{\phi}(\mathbf{x}, \mathbf{y})$ and $\mathbf{\mu}_{p}$, $\log \mathbf{\sigma}^2_{p}$ = $f_{\theta}(\mathbf{x})$.
        \STATE $\mathbf{\mu}_q'$ = $BN_{\gamma, \beta}(\mathbf{\mu}_q)$.
        \STATE Sample $\mathbf{z}\sim\mathcal{N}(\mathbf{\mu}'_q, \mathbf{\sigma}_q^2)$ and reconstruct $\mathbf{y}$ from $f_{\kappa}(\mathbf{z,x})$.
        \STATE Compute gradients $\mathbf{g}_{\phi,\theta,\kappa} \leftarrow \nabla_{\phi,\theta,\kappa}\mathcal{L}'$.
        \STATE Update $\phi, \theta,\kappa$ using $\mathbf{g}_{\phi,\theta,\kappa}$.
    \ENDFOR
    \end{algorithmic}
    \end{algorithm}
       
\begin{table*}[h]
\centering
\small
\begin{tabular*}{\textwidth}{c @{\extracolsep{\fill}} ccccccccc}
\hline
                                & \multicolumn{4}{c}{Yahoo}                                                                                                          & \multicolumn{4}{c}{Yelp}                                                                                                             \\ \hline 
\textbf{Model} & \textbf{NLL}         & \textbf{KL} & \textbf{MI} & \textbf{AU} & \textbf{NLL }    & \textbf{KL} & \textbf{MI} & \textbf{AU} \\ \hline
                                 & \multicolumn{8}{c}{Without a pretrained AE encoder}  \\
                                 CNN-VAE                & $\leq$332.1          & 10.0           & -           & -           & $\leq$359.1        & 7.6           & -           & -           \\
            LSTM-LM                & 328          & -           & -           & -           & 351.1        & -           & -           & -           \\
            VAE                             & 328.6                            & 0.0                          & 0.0                          & 0.0                          & 357.9                              & 0.0                          & 0.0                          & 0.0                          \\
% $\beta$-VAE (0.2)               & 332.2                            & 19.1                         & 3.3                          & 20.4                         & 360.7                              & 11.7                         & 3.0                          & 10.0                         \\
$\beta$-VAE (0.4)               & 328.7                           & 6.3                          & 2.8                          & 8.0                          & 358.2                              & 4.2                          & 2.0                          & 4.2                          \\
cyclic  $^*$                       & 330.6  & 2.1                          & 2.0                          & 2.3                         & 359.5                            & 2.0                          &1.9                         & 4.1                         \\
Skip-VAE $^*$                        & 328.5                            & 2.3                          & 1.3                          & 8.1                          & 357.6                             & 1.9                          & 1.0                          & 7.4                          \\
SA-VAE                          & 327.2                           & 5.2                          & 2.7                          & 9.8                          & \textbf{355.9} & 2.8                          & 1.7                          & 8.4                          \\
Agg-VAE                         & \textbf{326.7}  & 5.7                          & 2.9                          & 15.0                         & \textbf{355.9}                            & 3.8                          & 2.4                          & 11.3                         \\

FB (4)                         & 331.0  & 4.1                          & 3.8                          & 3.0                        & 359.2                            & 4.0                          &1.9                         & 32.0                         \\
FB (5)                         & 330.6  & 5.7                          & 2.0                          & 3.0                         & 359.8                            & 4.9                          &1.3                         & 32.0                         \\

$\delta$-VAE (0.1) $^*$                        & 330.7  & 3.2                          & 0.0                          & 0.0                         & 359.8  & 3.2                          & 0.0                          & 0.0                          \\
% $\delta$-VAE (0.15) $^*$                        & 331.6  & 4.8                          & 0.01                          & 0.0                         & 360.4  & 4.8                          & 0.04                          & 0.0                         \\
vMF-VAE (13)  $^*$                       & 327.4  & 2.0                         & -                          & 32.0                         & 357.5                           & 2.0                          & -                         & 32.0                         \\
BN-VAE (0.6) $^*$              & \textbf{326.7}                           & 6.2                          & 5.6                          & 32.0                        & 356.5                             & 6.5                         & 5.4                          & 32.0                         \\ 
BN-VAE (0.7) $^*$              & 327.4                           &   8.8                        & 7.4                          & 32.0                        & \textbf{355.9}                             & 9.1                          & 7.4                          & 32.0      
\\ 

\hline
& \multicolumn{8}{c}{With a pretrained AE encoder}                        \\
cyclic  $^*$                       & 333.1  & 25.8                         & 9.1                          & 32.0                         &  361.5                            & 20.5                          & 9.3                         & 32.0                         \\ 
FB (4)  $^*$                       & \textbf{326.2}  & 8.1                          & 6.8                          & 32.0                         & 356.0                            & 7.6                          &6.6                         & 32.0                         \\
% FB (5)  $^*$                       & 326.6  & 8.9                         & 7.3                          & 32.0                         & 356.5                            & 9.0                          & 7.4                          & 32.0                         \\ 
$\delta$-VAE (0.15)  $^*$                       & 331.0  & 5.6                         & 1.1                          & 11.2                         & 359.4                            & 5.2                          & 0.5                         & 5.9                         \\
vMF-VAE (13)  $^*$                       & 328.4  & 2.0                         & -                          & 32.0                         & 357.0                           & 2.0                          & -                         & 32.0                         \\
BN-VAE (0.6)  $^*$                       & 326.7  & 6.4                         & 5.8                          & 32.0                         & \textbf{355.5}                            & 6.6                          &5.9                         & 32.0                         \\
BN-VAE (0.7)  $^*$                       & 326.5  & 9.1                          & 7.6                          & 32.0                         & 355.7                            & 9.1                          & 7.5                          & 32.0                         \\ \hline
\end{tabular*}
\caption{Results on Yahoo and Yelp datasets. We report mean values across 5 different random runs. $^*$ indicates the results are from our experiments, while others are from \citet{he2019lagging, li-etal-2019-surprisingly}. We only show the best performance of every model for each dataset. More results on various parameters can be found in the Appendix. }\label{tb:text}
\end{table*} 
    
    \section{Experiments}
    % We test our approaches for preventing posterior collapse on three public benchmarks: two for language modeling (VAE) and one for dialogue generation (CVAE).
    %In each subsection, we first introduce the experiment setting and baseline models then we analyze the results in detail.

    %     \begin{table}[t]
    %     \centering
    %             \caption{Results of our model BN-VAE with different latent dimensions (z). $\gamma$ is the weight in BN.}
    %     \begin{tabular}{llllllll}
    %                     \hline
    %                     \multicolumn{2}{c}{\textbf{Params}} \multicolumn{3}{c}{\textbf{Yahoo}} & \multicolumn{3}{c}{\textbf{Yelp}} \\ \hline
    %                     \#$z$ & $\gamma$ &\textbf{NLL}  & \textbf{KL }   & \textbf{MI}   & \textbf{NLL}  & \textbf{KL}  & \textbf{MI}          \\ \hline
    %                     4  & 1.0 & 327.1    & 2.58   & 2.04  & 356.4   & 2.53  & 1.97      \\
    %                     8  & 0.75 & 327.2    & 2.76   & 2.14  & 356.4   & 2.81  & 2.13    \\
    %                     12 & 0.75 & 327.0   & 4.01       &  2.64     & \textbf{356.1 }   & 4.04  & 2.58        \\
    %                     16 & 0.75 & \textbf{326.8}    & 5.05   & 2.85  &     356.6     &    5.25   &   2.83          \\
    %                     32 & 0.5 & 327.4   & 4.15   & 2.51  &     357.1     &   4.04    &   2.41        \\ \hline
    %             \end{tabular}
    %             \label{tb:latensize}
    % \end{table}

    \subsection{VAE for Language Modeling}
    \label{sec:lm}
        {\bf Setup:}
    We test our approach on two benchmark datasets: Yelp and Yahoo corpora~\cite{yang2017improved}.
    % \footnote{***the dataset ref correct?}
    We use a Gaussian prior $\mathcal{N}(0, I)$, and the approximate posterior is a diagonal Gaussian. Following previous work \cite{burda2015importance,he2019lagging}, we report the estimated negative log likelihood (NLL) from 500 importance weighted samples, which can provide a tighter lower bound compared to the ELBO and shares the same information with the perplexity (PPL). Besides the NLL, we also report the KL, the mutual information (MI) $I_q$ \cite{alemi2016deep} and the number of activate units (AU)~\cite{burda2015importance} in the latent space. The $I_q$ can be calculated as:
        \begin{align}
    \label{eq:iq}
    I_q  = & \nonumber \mbox{E}_{p_d(\mathbf{x})}[KL(q_{\phi}(\mathbf{z}|\mathbf{x})||p(\mathbf{z}))] - \\
    & KL(q_{\phi}(\mathbf{z})||p(\mathbf{z})),
    \end{align}
    where $p_d(\mathbf{x})$ is the empirical distribution. The aggregated posterior $q_{\phi}(\mathbf{z}) = \mbox{E}_{p_d(\mathbf{x})}
[q_{\phi}(\mathbf{z}|\mathbf{x})]$ and $KL(q_{\phi}(\mathbf{z})||p(\mathbf{z}))$
can be approximated with Monte Carlo estimations.
 The AU is measured as $A_z = Cov(\mbox{E}_{\mathbf{z} \sim q(\mathbf{z}|\mathbf{x})}[\mathbf{z}])$. We set the threshold of 0.01, which means if $A_{zi} > 0.01$, the unit $i$ is active. 
    %Mutual information $I_q$ can be computed as in \cite{hoffman2016elbo} (eq. \ref{eq:iq}). It is actually a Monte Carlo estimate of an upper bound of mutual information \cite{dieng2018avoiding}.
    
    \noindent
    {\bf Configurations:} We use a 512-dimension word embedding layer for both datasets. For the encoder and the decoder, a single layer LSTM with 1024 hidden size is used. We use $\mathbf{z}$ to generate the initial state of the encoder following \citet{kim2018semi,he2019lagging,li-etal-2019-surprisingly}. To optimize the objective, we use mini-batch SGD with 32 samples per batch. We use one NVIDIA Tesla v100 for the experiments. For all experiments, we use the linear annealing strategy that increases the KL weight from 0 to 1 in the first 10 epochs if possible.
    
    \noindent
    {\bf Compared methods:}
    We compare our model with several strong baselines and methods that hold the previous state-of-the-art performance on text modeling benchmarks. 
    \vspace{-1.8mm}
    \begin{itemize}[wide=0\parindent,noitemsep]
        \item Baselines, including neural autoregressive models (the LSTM language model).
        \item Methods with weakening the decoder: CNN-VAE~\cite{yang2017improved}.
        \item Methods with a modified model structure: Skip-VAE \cite{dieng2018avoiding}.
        \item Methods with a modified training objective:
                    \begin{itemize}
                \item VAE with annealing \cite{bowman2015generating}.
                \item $\beta$-VAE \cite{higgins2017beta}.
                % which set a fixed KL weight.
                \item Cyclic annealing \cite{liu2019cyclical}, we use the default cyclic schedule.
            \end{itemize}
        \item Methods with a lower bound for KL values:
            \begin{itemize}
                \item Free-bits (FB) \cite{kingma2016improved}.
                % , we set $\lambda$ between 2, 3, 4, 5, 6, 7, 8.
                \item $\delta$-VAE \cite{razavi2018preventing}.
                \item vMF-VAE \cite{xu2018spherical}                % , we use $\delta$ between 0.1, 0.15, 0.2, 0.25, 0.3. 
                % \footnote{***i do not see that many parameter setting in the table? }
            \end{itemize}
        % \item Methods with modified training objective, including the VAE with annealing \cite{bowman2015generating}, $\beta$-VAE~\cite{higgins2017beta} which treats the KL weight as a hyperparameter and cyclic annealing \cite{liu2019cyclical} which changes the weight of the KL in a cyclic fashion. For $\beta$-VAE, we choose $\beta$ between 0.2, 0.4, 0.6 and 0.8. For cyclic annealing, we use the schedule in the original code.
        % \item Methods with a lower bound, including free-bits (FB) \cite{kingma2016improved} and $\delta$-VAE. We set $\lambda$ between 2, 3, 4, 5, 6, 7, 8 for FB and $\delta$ between 0.1, 0.15, 0.2, 0.25, 0.3 for $\delta$-VAE.
        % \item Methods with direct connections between $\mathbf{z}$ and $\mathbf{x}$, including Skip-VAE~\cite{dieng2018avoiding} which adds skip connections in the generative network.
        \item Methods with a modified training strategy. 
        \begin{itemize}
                \item Semi-amortized VAE (SA-VAE) \cite{kim2018semi}.
                \item VAE with an aggressive training (Agg-VAE) \cite{he2019lagging}.
                \item FB with a pretrained inference network (AE+FB) \cite{liu2019cyclical}
            \end{itemize}
        % including Semi-amortized VAE (SA-VAE)~\cite{kim2018semi} which composes the standard inference network with additional mean-field updates and Aggressive VAE (Agg-VAE)~\cite{he2019lagging} which performs an aggressively training to the inference network to reduce inference lag.
        % \item Free-bits with the pretrained encoder (AE+FB) \cite{li-etal-2019-surprisingly}. We experience with the same $\lambda$s with free-bits. We also add AE pretrained encoder to other methods to check whether it can further improve the performance.
        
    \end{itemize}
    %,the basic VAE, the VAE with annealing \cite{bowman2015generating}, ,   and . 
    % We also include several previous reported results \cite{yang2017improved}.  
    %We set $k=20, 80$ for the vMF-VAE.
    % We set $\gamma$ in our approach between 0.35, 0.45, 0.5, 0.55, 0.6, 0.65, 0.7, which is calculated by setting the lower bound of $E[KL]$ to 2, 3, 4, 5, 6, 7, 8.
    % \footnote{***move this para to the configuration?}
    % We vary different hyperparameters for the methods. More results can be found in the Appendix.
    \noindent
    {\bf Main results:}
    Table \ref{tb:text} shows the results. 
     We further split the results into two different settings, one for models with a pretrained inference network and one without it. Our approach achieves the best NLL in the setting without a pretrained inference network on both datasets and is competitive in the setting with a pretrained encoder.
     %\footnote{***u did not state ur settings; do u want to specify the row number for each observation below?}
     Moreover, we can observe that:
     \vspace{-1.8mm}
     \begin{itemize}[wide=0\parindent,noitemsep]
        %  \item Without annealing, SA-VAE fails to maintain the KL positive. Due to the existence of a fixed positive lower bound, our approach can avoid posterior collapse in both settings.
         \item $\delta$-VAE does not perform well in both settings, which shows that constraining the parameters in a small interval is harmful to the model. In vMF-VAE, data points share the same KL value. Our approach is flexible and gets better performance.
        %  \item Free-bits along does not provide a superior performance. When combined with a pretrained inference network, it works quite well.   
         \item Although Agg-VAE and SA-VAE both get good performance, they require additional updates on the inference network and cost more training efforts, which are validated in the next part.
         \item Cyclic annealing with a pretrained inference network achieves the highest KL, but it may not be a good generative model.
         \item Paired with a pretrained inference network, all methods except cyclic annealing can someway boost the performance. This phenomenon indicates that the lagging problem \cite{he2019lagging} is important in VAE training. When leveraging the pretrained inference network, our approach achieves the smallest performance gap compared with other methods. In other words, our approach can alleviate the lagging problem efficiently.
     \end{itemize}    

\begin{table}[t]
\centering
\small
\begin{tabular}{lcccc} \hline
                & \multicolumn{2}{c}{\textbf{Yahoo}} & \multicolumn{2}{c}{\textbf{Yelp}}  \\ \cline{2-5}
                \textbf{Model} & \textbf{Hours}      & \textbf{Ratio}      & \textbf{Hours}      & \textbf{Ratio}           \\ \hline
                VAE       &     3.83          &    1.00   &    4.50       &     1.00          \\
                SA-VAE       &     52.99          &    12.80   &     59.37      &     12.64          \\
                Agg VAE   &   11.76  &    2.84         &        21.44       &    4.56          \\
                AE+FB   &   7.70  &    2.01         &        9.22       &    2.05 \\
                BN-VAE &         3.98      &    1.04      &       4.60       &     1.02         \\ \hline     
        \end{tabular}
                \caption{Comparison of training time to convergence. We report both the absolute hours and relative speed. }    \label{tb:time}
\end{table}
      \noindent
    % {\bf Varying the latent size:} 
    % We notice that it is hard to find the relationship between the number of AU and the performance, e.g., SA-VAE and Agg-VAE both get 355.9 in terms of the NLL on Yelp, but with different numbers of AU. Due to the regularization in our approach, the number of AU is always the same as the latent size in BN-VAE. To study how the number of AU affects the performance, we also report the performance with various latent sizes in Table~\ref{tb:latensize}. The best performance is achieved when the latent size is 16 for Yahoo and 12 for Yelp. Thus, we conjecture that datasets with richer information may need more latent variables to encode necessary information. It is an interesting direction to choose a proper number of latent dimension automatically in our future work. 
    
    % Besides the latent size, we prefer a large $\gamma$ due to that a small $\gamma$ will force the latent values to converge to the mean value which is harmful to the model.   %\footnote{***add more analysis to show the adv of our method. always say the most important results/analysis first.}

         \noindent
    {\bf Training time:}
    Table~\ref{tb:time} shows the training time (until convergence) and the relative ratio of the basic VAE, our approach and the other best three models in Table~\ref{tb:text}. SA-VAE is about 12 times slower than our approach due to the local update for each data point. Agg-VAE is 2-4 times slower than ours because it requires additional training for the inference network. AE+FB needs to train an autoencoder before the VAE. However, our approach is fast since we only add one-layer batch normalization, and thus the training cost is almost the same as the basic VAE. More results about the training behavior can be found in Section 3 of the Appendix.

\begin{table}[]
\centering
\small
%\begin{adjustbox}{max width=0.46\textwidth}
\begin{tabular}{lccccc}
\hline
\textbf{\#label}     & \textbf{100}   & \textbf{500}   & \textbf{1k}  & \textbf{2k} & \textbf{10k} \\ \hline
AE           & 81.1  & 86.2  & 90.3  & 89.4 & 94.1  \\ 
VAE          & 66.1  & 82.6  & 88.4  & 89.6 & 94.5  \\ 
$\delta$-VAE & 61.8  &  61.9 & 62.6  & 62.9 & 93.8     \\
Agg-VAE      & 80.9  & 85.9  & 88.8  & 90.6 & 93.7  \\ 
cyclic        & 62.4  & 75.5  & 80.3  & 88.7 & 94.2  \\ 
FB (9)       & 79.8  & 84.4  & 88.8  & 91.12 & 94.7  \\ 
AE+FB (6)     & 87.6  & 90.2  & 92.0  & 93.4 & 94.9  \\  \hline
BN-VAE (0.7) & \textbf{88.8} & \textbf{91.6} & \textbf{92.5} & \textbf{94.1} & \textbf{95.4} \\\hline
\end{tabular}
%\end{adjustbox}
\caption{Accuracy on Yelp.}
\label{tb:classificaion}
\end{table}

\begin{table}[]
\centering
\small
\begin{tabular}{lccc}
\hline
\textbf{Model} & \textbf{CVAE}  & \textbf{CVAE (BOW)} & \textbf{BN-VAE} \\ \hline
PPL   & 36.40 & 24.49    & 30.67  \\ \hline
KL    & 0.15  & 9.30     & 5.18   \\ \hline
BLEU-4   & 10.23 & 8.56     & 8.64   \\ \hline
A-bow Prec   & 95.87 & 96.89    & 96.64  \\ \hline 
A-bow Recall   & 90.93 & 93.95    & 94.43  \\\hline
E-bow Prec   & 86.26 & 83.55    & 84.69  \\\hline
E-bow Recall   & 77.91 & 81.13    & 81.75  \\\hline
\end{tabular}
\caption{Comparison on dialogue generation. }\label{tb:cvaes}
\end{table}    

\begin{table*}[h]
        \centering
        \small
            \begin{tabular}{lccc|ccc|ccc} \hline
 & \multicolumn{3}{c}{\textbf{Fluency}}                                                                & \multicolumn{3}{c}{\textbf{Relevance}}                                                             & \multicolumn{3}{c}{\textbf{Informativeness}}                                                      \\ \cline{2-10}
            \textbf{Model} & \multicolumn{1}{c}{\textbf{Avg}}               & \multicolumn{1}{c}{\textbf{\#Accept}} & \multicolumn{1}{c|}{\textbf{\#High}} & \multicolumn{1}{c}{\textbf{Avg}}               & \multicolumn{1}{c}{\textbf{\#Accept}} & \multicolumn{1}{c|}{\textbf{\#High}} & \multicolumn{1}{c}{\textbf{Avg}}               & \multicolumn{1}{c}{\textbf{\#Accept}} & \multicolumn{1}{c}{\textbf{\#High}} \\ \hline
CVAE                            & 2.11 (0.58)                           & 87\%                    & 23\%                     & 1.90 (0.49)                           & 82\%                    & 8\%                      & 1.39 (0.59)                           & 34\%                    & 5\%                     \\ \hline
CVAE (BOW)                      & 2.08 (0.73)                           & 84\%                    & 23\%                     & 1.86 (0.58)                           & 75\%                    & 11\%                     & \textbf{1.54} (0.65) & 46\%                    & 8\%                     \\ \hline
BN-CVAE                         & \textbf{2.16} (0.71) & 88\%                    & 27\%                     & \textbf{1.92} (0.67) & 80\%                    & 12\%                     & \textbf{1.54} (0.67) & 43\%                    & 10\%                    \\ \hline
\end{tabular}
%}
\caption{Human evaluation results. Numbers in parentheses is the corresponding variance on 200 test samples.} \label{tb:human}
\end{table*}

\begin{table*}[h]
        \centering
        \small
    \begin{tabular}{llll}
        \hline
        \multicolumn{4}{l}{Topic: ETHICS IN GOVERNMENT}                                                                                                                              \\ \hline
        \multicolumn{4}{l}{Context: have trouble drawing lines as to what's illegal and what's not}                                                                              \\ \hline
        \multicolumn{4}{l}{Target (statement): well i mean the other problem is that they're always up for}                                                                          \\ \hline
        \multicolumn{2}{l|}{CVAE}    & \multicolumn{1}{l|}{CVAE (BOW)}             & \multicolumn{1}{l}{BN-CVAE}                                                                   \\ 
        \multicolumn{2}{l|}{1. yeah} & \multicolumn{1}{l|}{1. yeah}                & \multicolumn{1}{l}{1. it's not a country}                                                       \\ 
        \multicolumn{2}{l|}{2. yeah} & \multicolumn{1}{l|}{2. oh yeah they're not} & \multicolumn{1}{l}{2. it is the same thing that's what i think is about the state is a state} \\ 
        \multicolumn{2}{l|}{3. yeah} & \multicolumn{1}{l|}{3. no it's not too bad} & \multicolumn{1}{l}{3. yeah it's}                                                              \\ \hline
    \end{tabular}
    \caption{Sampled generated responses. Only the last sentence in the context is shown here.}
    \label{tb:example}
\end{table*} 

\noindent
    {\bf Performance on a downstream task - Text classification:}
    The goal of VAE is to learn a good representation of the data for downstream tasks. Here, we evaluate the quality of latent representations by training a one-layer linear classifier based on the mean of the posterior distribution. We use a downsampled version of the Yelp sentiment dataset \cite{shen2017style}. \citet{li-etal-2019-surprisingly} further sampled various labeled data to train the classifier. To compare with them fairly, we use the same samples in \citet{li-etal-2019-surprisingly}. Results are shown in Table \ref{tb:classificaion}. Our approach achieves the best accuracy in all the settings. For 10k training samples, all the methods get a good result. However, when only using 100 training samples, different methods vary a lot in accuracy. The text classification task shows that our approach can learn a good latent representation even without a pretrained inference network. 

    \subsection{CVAE for Dialogue Generation}
      \noindent
    {\bf Setup:}
    For dialogue generation, we test our approach in the setting of CVAE. Following previous work~\cite{zhao2017learning}, we use the Switchboard (SW) Corpus \cite{godfrey1997switchboard}, which contains 2400 two-sided telephone conversations. We use a bidirectional GRU with hidden size 300 to encode each utterance and then a one-layer GRU with hidden size 600 to encode previous $k$-1 utterances as the context. The response decoder is a one-layer GRU with hidden size 400. The latent representation $\mathbf{z}$ has a size of 200. We use the evaluation metrics from \citet{zhao2017learning}: (1) Smoothed Sentence-level BLEU~\cite{chen2014systematic}; (2) Cosine Distance of Bag-of-word Embedding, which is a simple method to obtain sentence embeddings. We use the pretrained Glove embedding~\cite{pennington2014glove} and denote the average method as A-\textit{bow} and the extreme method as E-\textit{bow}. Higher values indicate more plausible responses. We compared our approach with CVAE and CVAE with bag-of-words (BOW) loss~\cite{zhao2017learning},
    which requires the decoder in the generation network
to predict the bag-of-words in the response $\textbf{y}$ based on $\mathbf{z}$.
    
      \noindent
    {\bf Automatic evaluation:}
    Table~\ref{tb:cvaes} shows the results of these three approaches. From the KL values, we find that CVAE suffers from posterior collapse while CVAE (BOW)  and our approach avoid it effectively. For BLEU-4, we observe the same phenomenon in the previous work~\cite{liu2019cyclical,zhao2017learning} that CVAE is slightly better than the others. This is because CVAE tends to generate the most likely and safe responses repeatedly with the collapsed posterior. As for precision, these three models do not differ much. However, CVAE (BOW) and our BN-VAE outperform CVAE in recall with a large margin. This indicates that BN-VAE can also produce diverse responses with good quality like CVAE (BOW).
    
      \noindent
    {\bf Human evaluation:}
    We conduct the human evaluation by asking five annotators from a commercial annotation company to grade 200 sampled conversations from the aspect of fluency, relevance and informativeness on a scale of 1-3 (see Section 4 of the Appendix for more details on the criteria).
    We also report the proportion of acceptable/high scores ($\geq 2$ and $=3$) on each metric.
    Table \ref{tb:human} shows 
    the annotation results. Overall, our approach beats the other two compared methods in relevance and fluency with more informative responses. Also, our approach has the largest proportion of responses whose scores are \textit{High}. This indicates that our model can produce more meaningful and relevant responses than the other two.
    
      \noindent
    {\bf Case study:}
    Table \ref{tb:example} shows the sampled responses generated by the three methods (more can be found in the Appendix). By maintaining a reasonable KL, responses generated by our approach are more relevant to the query with better diversity compared to the other two. We test the three methods in the simplest setting of dialogue generation. Note that the focus of this work is to improve the CVAE itself by avoiding its KL vanishing problem but not to hack the state-of-the-art dialogue generation performance. To further improve the quality of generated responses, we can enhance our approach by incorporating knowledge such as dialogue acts~\cite{zhao2017learning}, external facts~\cite{ghazvininejad2018knowledge} and personal profiles~\cite{zhang-etal-2018-personalizing}.

    \section{Conclusions and Future Work}
    
    In this paper, we tackle the posterior collapse problem when VAE is paired with autoregressive decoders. Instead of considering the KL individually, we make it follow a distribution $D_{KL}$ and show that keeping the expectation of $D_{KL}$ positive is sufficient to prevent posterior collapse. We propose Batch Normalized VAE (BN-VAE), a simple but effective approach to set a lower bound of $D_{KL}$ by regularization the approximate posterior's parameters. Our approach can also avoid the recently proposed lagging problem efficiently without additional training efforts. We show that our approach can be easily extended to CVAE. We test our approach on three real applications, language modeling, text classification and dialogue generation. Experiments show that our approach outperforms strong baselines and is competitive with more complex methods which keeping substantially faster.

    We leverage the Gaussian prior as the example to introduce our method in this work. The key to our approach to be applicable is that we can get a formula for the expectation of the KL. However, it is hard to get the same formula for some more strong or sophisticated priors, e.g., the Dirichlet prior. For these distributions, we can approximate them by the Gaussian distributions (such as in \citet{srivastava2017autoencoding}). In this way, we can batch normalize the corresponding parameters. Further study in this direction may be interesting.

\bibliographystyle{acl_natbib}
\bibliography{acl2020}

\newpage

\appendix

\section{Appendix}
\label{sec:appendix}

    \begin{figure*}
        \centering
        \includegraphics[width=\textwidth]{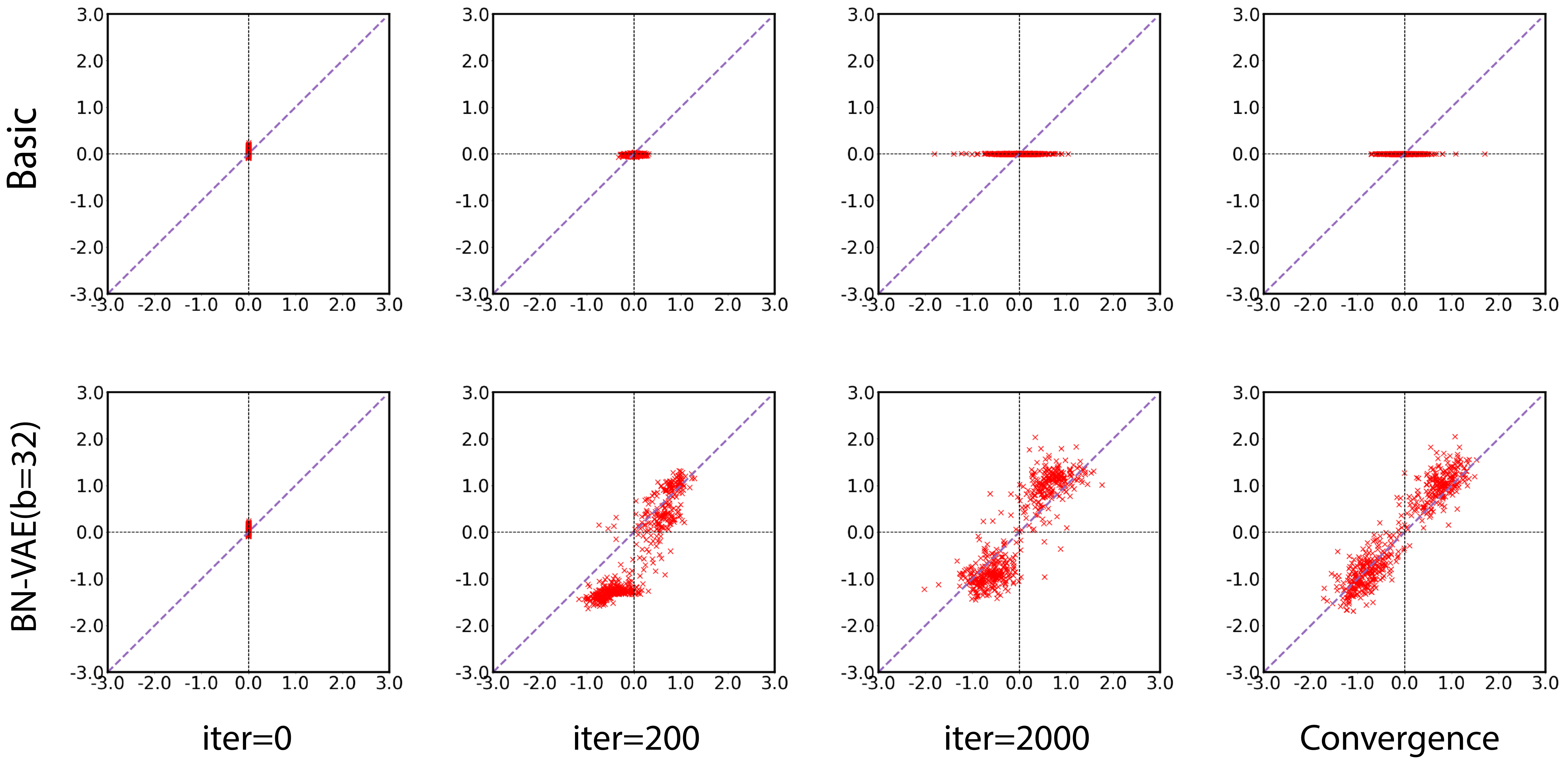}
        \caption{Visualization of 500 sampled data from the synthetic dataset during the training. The x-axis is $\mu_{x,\theta}$, the approximate model posterior mean. The y-axis is $\mu_{x,\phi}$, which represents the inference posterior mean. b is batch size and $\gamma$ is 1 in BN.}
        \label{fig:synthetic1}
    \end{figure*}

\begin{figure*}[]
\centering
\includegraphics[width=\textwidth]{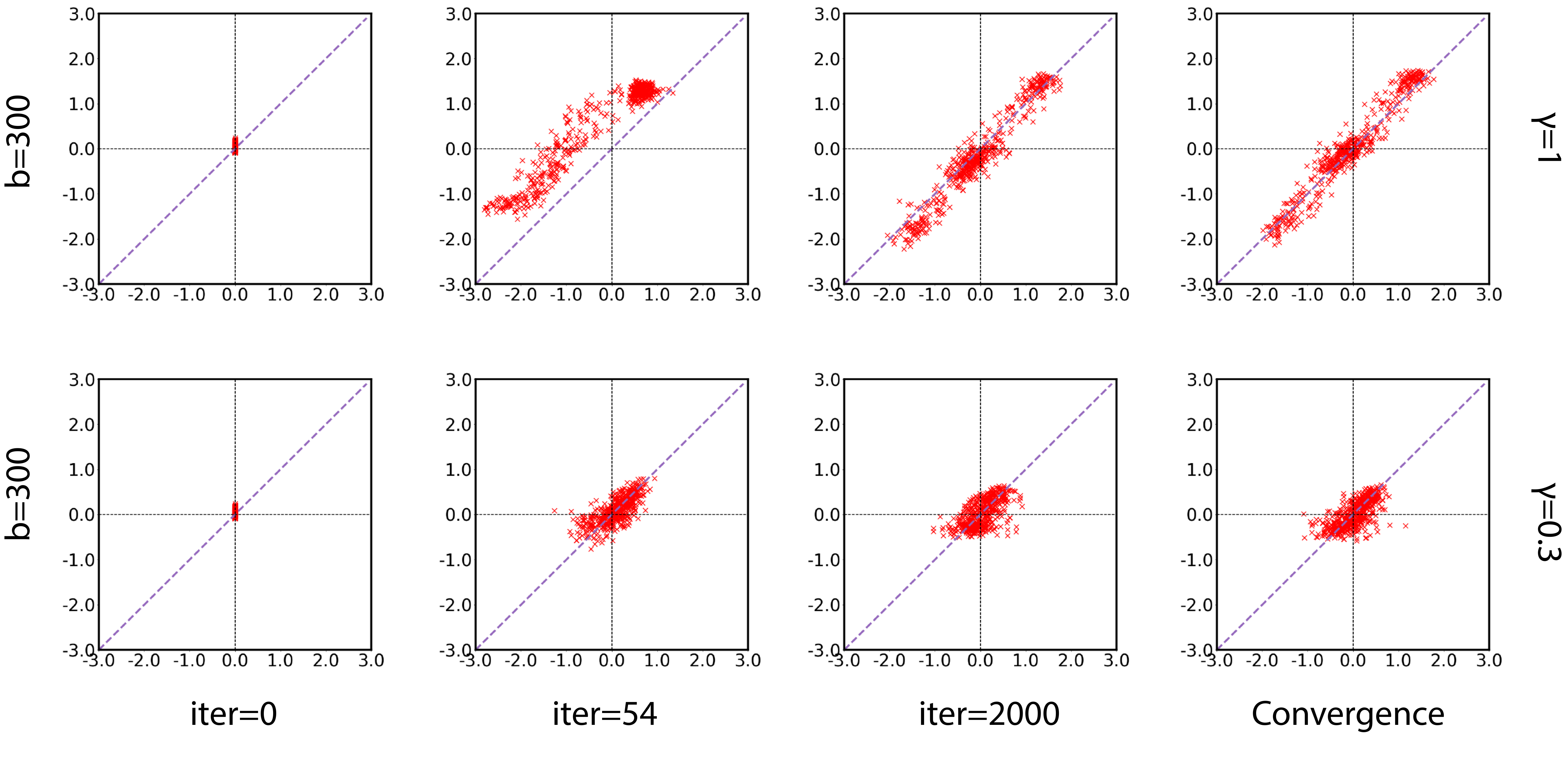}
\caption{Visualization of our BN-VAE on different $\gamma$ for synthetic data.}
\label{fig:synthetic}
\end{figure*}

\begin{figure*}[]
	\centering
	\includegraphics[width=1.8\columnwidth]{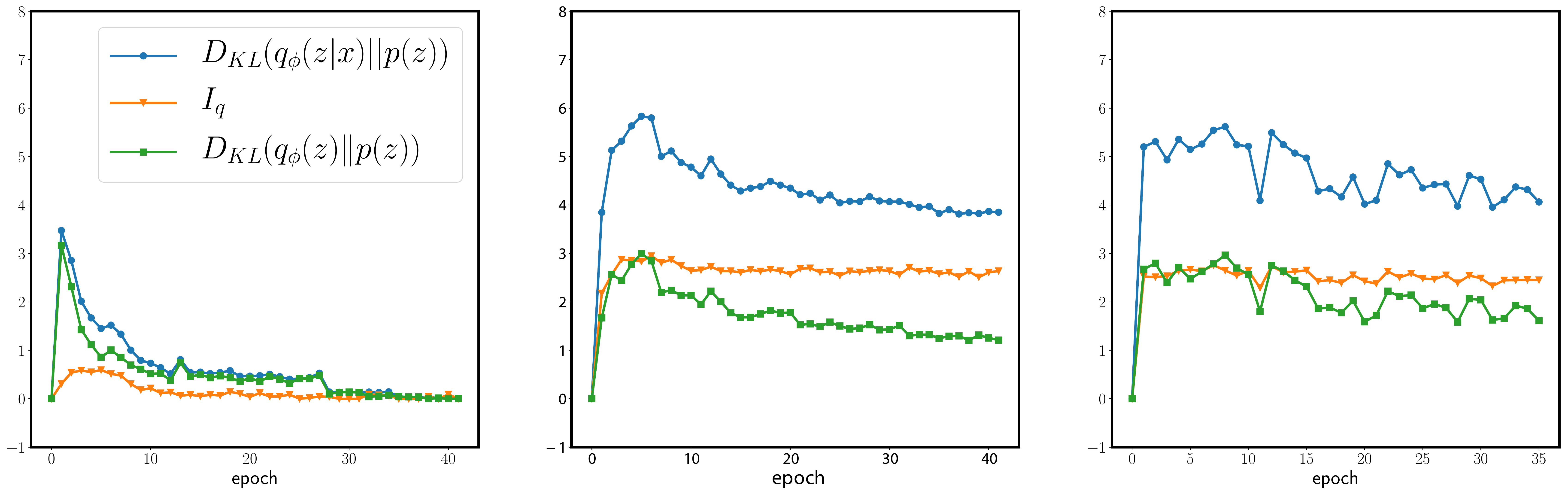}
	\caption{Training behavior on Yelp. Left/Middle/Right: VAE/Agg-VAE/BN-VAE (all models are with annealing).}
	\label{fig:kl}
\end{figure*}

\begin{table*}[]
\centering
\caption{Human evaluation criteria.}
 \setlength{\tabcolsep}{0.5mm}
\begin{tabular}{c|l|l|l}
\hline
\multicolumn{1}{l}{}  & \multicolumn{1}{c}{Fluency}                                                                                                                      & \multicolumn{1}{c}{Relevance}                                                                                                                     & \multicolumn{1}{c}{Informativeness}                                                                        \\ \hline
1 Point                     & \begin{tabular}[c]{@{}l@{}}1. Hard to understand\\ 2. Too many syntax mistakes\end{tabular}                                                      & Not related to the query at all                                                    & \begin{tabular}[c]{@{}l@{}}1. Generic responses.\\ 2. Repeated query.\end{tabular}                         \\ \hline
2 Points                    & \begin{tabular}[c]{@{}l@{}}1. Several syntax mistakes but \\ still understandable\\ 2. short responses, e.g., Generic \\ responses\end{tabular} & \begin{tabular}[c]{@{}l@{}}1. Response and query are in\\  the same domain/topic but \\ are not directly related\\ 2. Generic responses\end{tabular} & between 1 and 3.                                                                                           \\ \hline
\multicolumn{1}{l|}{3 Points} & \begin{tabular}[c]{@{}l@{}}Only few syntax mistakes with \\ a moderate length\end{tabular}                                      & closely related to the query                                                      & \begin{tabular}[c]{@{}l@{}}1. Creative responses. \\ 2. Contain new information \\ about the query.\end{tabular} \\ \hline
\end{tabular}
\label{tb:evaluate}
\end{table*}

\subsection{Experiments on Synthetic Data }

We follow the Agg-VAE and construct the synthetic data to validate whether our approach can avoid the lagging problem. VAE used in this synthetic task has a LSTM encoder and a LSTM decoder. We use a scalar latent variable because we need to compute $\mu_{x,\theta}$ which is approximated by discretization of $p_{\theta}(z|x)$. To visualize the training progress, we sample 500 data points from the validation set and show them on the mean space.

We plot the mean value of the approximate posterior and the model posterior during training for the basic VAE and BN-VAE. As shown the first column in Fig.~\ref{fig:synthetic1}, all points have the zero mean of the model posterior (the x-axis), which indicates that $\mathbf{z}$ and $\mathbf{x}$ are independent at the beginning of training. For the basic VAE, points start to spread in the x-axis during training while sharing almost the same y value, since the model posterior $p_\theta(\mathbf{z}|\mathbf{x})$ is well learned with the help of the autoregressive decoder. However, the inference posterior $q_\phi(\mathbf{z}|\mathbf{x})$ is lagging behind $p_\theta(\mathbf{z}|\mathbf{x})$ and collapses to the prior in the end. Our regularization approximated by BN, on the other hand, pushes the inference posterior $q_{\phi}(\mathbf{z}|\mathbf{x})$ away from the prior ($p(\mathbf{z})$) at the initial training stage, and forces $q_\phi(\mathbf{z}|\mathbf{x})$ to catch up with $p_\theta(\mathbf{z}|\mathbf{x})$ to minimize $KL(q_{\phi}(\mathbf{z}|\mathbf{x})||p_{\theta}(\mathbf{z}|\mathbf{x}))$ in Eq. 9. As in the second row of Fig.~\ref{fig:synthetic1}, points spread in both directions and towards the diagonal. 

We also report the results on different $\gamma$'s with different batch sizes (32 in Fig. \ref{fig:synthetic1}). Fig.~\ref{fig:synthetic} shows the training dynamics. Both settings of $\gamma$ avoid posterior collapse efficiently. A larger $\gamma$ produces more diverse $\mu$'s which spread on the diagonal. However, a small $\gamma$ results in a small variance for the distribution of $\mu$, thus $\mu$'s in the bottom row are closer to the original (mean of the distribution). When $\gamma$ is 0, posterior collapse happens. Different batch sizes do not diff a lot, so 32 is a decent choice. An intuitive improvement of our method is to automatically learn different $\gamma$ for different latent dimensions, which we leave for future work. 

\subsection{Proof in CVAE} 
The KL can be computed as:
\begin{align}
        KL &=\frac{1}{2}\sum_{i=1}^n(\frac{\sigma_{qi}^2 
         + (\mu_{qi}-\mu_{pi})^2}{\sigma_{pi}^2} \\ \nonumber
         &+\sigma_{pi}^2+\mu_{pi}^2-log\sigma_{qi}^2-1).
\end{align}
We need to prove that KL will not achieve the minimum number when $\mu_{pi}$ equals to $\mu_{qi}$ and $\sigma_{pi}$ equals $\sigma_{qi}$. We take hidden size as 1 for example. The binary function about $\mu_{pi}$ and $\sigma_{pi}$ is:
\begin{align}
    f_{\mu_{pi},\sigma_{pi}} &= (\frac{\sigma_{qi}^2 + (\mu_{qi}-\mu_{pi})^2}{\sigma_{pi}^2}\\ \nonumber
    &+\sigma_{pi}^2+\mu_{pi}^2-log\sigma_{qi}^2-1),
\end{align}
the maxima and minima of $f_{\mu_{pi},\sigma_{pi}}$ must be the stationary point of $f_{\mu_{pi},\sigma_{pi}}$ due to its continuity. The stationary point is:
\begin{align}
    \frac{\partial f}{\partial \mu_{pi}} =\frac{2(\mu_{pi} - \mu_{qi})}{\sigma_{pi}^2} + 2\mu_{pi} 
\end{align}
\begin{align}
    \frac{\partial f}{\partial \sigma_{pi}} =\frac{-2(\sigma_{qi}^2 + (\mu_{qi}-\mu_{pi})^2)}{\sigma_{pi}^3} + 2\sigma_{pi}.
\end{align}
When $\mu_{pi}=\mu_{qi}$ and $\sigma_{pi}=\sigma_{qi}$, both partial derivative is not 0. So it is not the stationary point of $f$, then it won't be the minima.

\subsection{Language Modeling}

We investigate the training procedure for different models. We plot the MI $I_q$, $D_{KL}$ in the ELBO and the distance between the approximated posterior and the prior, $D_{KL}(q_{\phi}(z)||p(z))$. As in Eq. 4 in the main paper,  $D_{KL}$ in the ELBO is the sum of the other two. Fig.~\ref{fig:kl} shows these three values throughout the training. Although $D_{KL}$ is the upper bound of the mutual information, we notice that the gap is usually large. In the initial training stage, $D_{KL}$ increases in the basic VAE with annealing, while its MI remains small. With the weight decreases, the method finally suffers from posterior collapse. In contrast, our approach can obtain a high MI with a small $D_{KL}$ value like aggressive VAE. The full results on language modeling are in Table \ref{tb:text2}.
\begin{table*}[h]
\centering
\small
\begin{tabular*}{\textwidth}{l @{\extracolsep{\fill}} ccccccccc}
\hline
                                & \multicolumn{4}{c}{Yahoo}                                                                                                          & \multicolumn{4}{c}{Yelp}                                                                                                             \\ \hline 
\textbf{Model} & \textbf{NLL}         & \textbf{KL} & \textbf{MI} & \textbf{AU} & \textbf{NLL }    & \textbf{KL} & \textbf{MI} & \textbf{AU} \\ \hline
                                 CNN-VAE                & $\leq$332.1          & 10.0           & -           & -           & $\leq$359.1        & 7.6           & -           & -           \\
            LSTM-LM                & 328          & -           & -           & -           & 351.1        & -           & -           & -           \\
            VAE                             & 328.6                            & 0.0                          & 0.0                          & 0.0                          & 357.9                              & 0.0                          & 0.0                          & 0.0                          \\
$\beta$-VAE (0.2)               & 332.2                            & 19.1                         & 3.3                          & 20.4                         & 360.7                              & 11.7                         & 3.0                          & 10.0                         \\
$\beta$-VAE (0.4)               & 328.7                           & 6.3                          & 2.8                          & 8.0                          & 358.2                              & 4.2                          & 2.0                          & 4.2                          \\
$\beta$-VAE (0.6)               & 328.5                           & 0.3                          & 0.0                          & 1.0                          & 357.9                              & 0.2                         & 0.1                          & 3.8                          \\
$\beta$-VAE (0.8)               & 328.8                           & 0.0                          & 0.0                          & 0.0                          & 358.1                              & 0.0                          & 0.0                         & 0.0                          \\
cyclic  $^*$                       & 330.6  & 2.1                          & 2.0                          & 2.3                         & 359.5                            & 2.0                          &1.9                         & 4.1                         \\
Skip-VAE $^*$                        & 328.5                            & 2.3                          & 1.3                          & 8.1                          & 357.6                             & 1.9                          & 1.0                          & 7.4                          \\
SA-VAE                          & 327.2                           & 5.2                          & 2.7                          & 9.8                          & \textbf{355.9} & 2.8                          & 1.7                          & 8.4                          \\
Agg-VAE                         & \textbf{326.7}  & 5.7                          & 2.9                          & 15.0                         & \textbf{355.9}                            & 3.8                          & 2.4                          & 11.3                         \\

FB (4)                         & 331.0  & 4.1                          & 3.8                          & 3.0                        & 359.2                            & 4.0                          &1.9                         & 32.0                         \\
FB (5)                         & 330.6  & 5.7                          & 2.0                          & 3.0                         & 359.8                            & 4.9                          &1.3                         & 32.0                         \\

$\delta$-VAE (0.1) $^*$                        & 330.7  & 3.2                          & 0.0                         & 0.0                         & 359.8  & 3.2                          & 0.0                          & 0.0                          \\
$\delta$-VAE (0.15) $^*$                        & 331.6  & 4.8                          & 0.0                        & 0.0                         & 360.4  & 4.8                          & 0.0                         & 0.0                         \\
$\delta$-VAE (0.2) $^*$                        & 332.2  & 6.4                          & 0.0                         & 0.0                         & 361.5  & 6.4                          & 0.0                         & 0.0                          \\
$\delta$-VAE (0.25) $^*$                        & 333.5  & 8.0                          & 0.0                          & 0.0                         & 362.5  & 8.0                          & 0.0                          & 0.0                         \\
vMF-VAE (13) $^*$                        & 327.4  & 2.0                         & -                         & 32.0                        & 357.5  & 2.0                          & -                          & 32.0                          \\
vMF-VAE (16) $^*$                        & 328.5  & 3.0                          & -                        & 32.0                         & 367.8  & 3.0                          & -                         & 32.0                         \\
vMF-VAE (20) $^*$                        & 329.4  & 4.0                          & --                        & 32.0                         & 358.0  & 4.0                          & -                         & 32.0                          \\
vMF-VAE (23) $^*$                        & 328.7  & 5.0                          & -                         & 32.0                         & 357.3  & 5.0                          & -                          & 32.0                         \\
vMF-VAE (25) $^*$                        & 330.1  & 6.0                          & -                         & 32.0                         & 357.8  & 6.0                          & -                         & 32.0                          \\
vMF-VAE (30) $^*$                        & 329.5  & 7.0                          & -                          & 32.0                         & 357.8  & 7.0                          & -                          & 32.0                         \\
BN-VAE (0.3) $^*$              & 328.1                          & 1.6                          & 1.4                          & 32.0                        & 356.7                             & 1.7                        & 1.4                          & 32.0                         \\ 

BN-VAE (0.4) $^*$              & 327.7                          & 2.7                          & 2.2                          & 32.0                        & 356.2                             & 3.1                         & 2.5                          & 32.0                         \\ 
BN-VAE (0.5) $^*$              & 327.4                           &   4.2                        & 3.3                         & 32.0                        & 356.4                             & 4.4                          & 3.8                          & 32.0                         \\ 
BN-VAE (0.6) $^*$              & \textbf{326.7}                           & 6.2                          & 5.6                          & 32.0                        & 356.5                             & 6.5                         & 5.4                          & 32.0                         \\ 
BN-VAE (0.7) $^*$              & 327.4                           &   8.8                        & 7.4                          & 32.0                        & \textbf{355.9}                             & 9.1                          & 7.4                          & 32.0                         \\ 

Pretrained encoder \\

 +cyclic  $^*$                       & 333.1  & 25.8                         & 9.1                          & 32.0                         &  361.5                            & 20.5                          & 9.3                         & 32.0                         \\ 
  +FB (2)  $^*$                       & 327.2  & 4.3                          & 3.8                          & 32.0                         & 356.6                            & 4.6                          & 4.2                        & 32.0                         \\
 +FB (3)  $^*$                       & 327.1 & 4.5                         & 3.9                         & 32.0                         & 356.3                            & 5.8                          & 5.2                          & 32.0                         \\ 
  +FB (4)  $^*$                       & \textbf{326.2}  & 8.1                          & 6.8                          & 32.0                         & 356.0                            & 7.6                          &6.6                         & 32.0                         \\
 +FB (5)  $^*$                       & 326.6  & 8.9                         & 7.3                          & 32.0                         & 356.5                            & 9.0                          & 7.4                          & 32.0                         \\ 
 +FB (6)  $^*$                       & 326.6  & 10.8                          & 8.1                          & 32.0                         & 356.5                           & 12.0                          &8.6                         & 32.0                         \\
 +FB (7)  $^*$                       & 326.6  & 12.1                         & 8.5                         & 32.0                         & 356.8                            & 13.4                         & 8.9                         & 32.0                         \\ 
  +FB (8)  $^*$                       & 326.7  & 13.6                         & 8.9                          & 32.0                         & 357.5                           & 15.8                         &9.2                        & 32.0                         \\

 +$\delta$-VAE (0.15)  $^*$                       & 331.0  & 5.6                         & 1.1                          & 11.2                         & 359.4                            & 5.2                          & 0.5                         & 5.9                         \\
 vMF-VAE (13)  $^*$                       & 328.4  & 2.0                         & -                          & 32.0                         & 357.0                           & 2.0                          & -                         & 32.0                         \\
 +BN-VAE (0.6)  $^*$                       & 326.7  & 6.4                         & 5.8                          & 32.0                         & \textbf{355.5}                            & 6.6                          &5.9                         & 32.0                         \\
 +BN-VAE (0.7)  $^*$                       & 326.5  & 9.1                          & 7.6                          & 32.0                         & 355.7                            & 9.1                          & 7.5                          & 32.0                         \\ \hline
\end{tabular*}
\caption{Results on Yahoo and Yelp datasets. We report mean values across 5 different random runs. $^*$ indicates the results are from our experiments, while others are from previous report.}\label{tb:text2}
\end{table*} 

\subsection{CVAE for dialogue generation}

{\bf Human evaluation:} We evaluate the generated responses from three aspects: relevance, fluency and informativeness. Here we introduce the criteria of the evaluation as shown in Table \ref{tb:evaluate}. We sample 200 conversations from the test set. For each conversation, we sample three generated responses from each model, totally 600 responses.

\noindent
{\bf Case study:} We report 4 examples generated from these three models, shown in Table~\ref{tb:example2}. CVAE (BOW) and our approach both can generate diverse responses. However, responses from ours are more related to the context compared with the other two.

\begin{table*}
\centering
\caption{Sampled generated responses. Only the last sentence in the context is shown here.}
\label{tb:example2}
\begin{tabular}{l|l|l} 
\hline
\multicolumn{3}{l}{Topic: ETHICS IN GOVERNMENT}                                                                                                                                                                                                                                                                                              \\ 
\hline
\multicolumn{3}{l}{Context: have trouble drawing lines as to what's illegal and what's not}                                                                                                                                                                                                                                                 \\ 
\hline
\multicolumn{3}{l}{Target (statement): well i mean the other problem is that they'are always up for}                                                                                                                                                                                                                                        \\ 
\hline
\multicolumn{1}{l|}{CVAE}    & CVAE (BOW)                                                                                                                                                       & \multicolumn{1}{l}{BN-CVAE}                                                                                                               \\ 

\multicolumn{1}{l|}{1. yeah} & 1. yeah                                                                                                                                                          & \multicolumn{1}{l}{1. it's not a country}                                                                                                 \\ 

\multicolumn{1}{l|}{2. yeah} & 2. oh yeah, they're not                                                                                                                                          & \multicolumn{1}{l}{\begin{tabular}[c]{@{}l@{}}2. it is the same thing that's~\\what i think is about the state\\is a state\end{tabular}}  \\ 

\multicolumn{1}{l|}{3. yeah} & 3. no it's not too bad                                                                                                                                           & \multicolumn{1}{l}{3. yeah it's}                                                                                                          \\ 
\hline
\multicolumn{3}{l}{Topic:VACATION SPOTS}                                                                                                                                                                                                                                                                                                      \\ 
\hline
\multicolumn{3}{l}{Context: well i ' ll talk to you later}                                                                                                                                                                                                                                                                                    \\ 
\hline
\multicolumn{3}{l}{Target (conventional-closing) : okay now do you push the buttons now}                                                                                                                                                                                                                                                      \\ 
\hline
CVAE                          & CVAE (BOW)                                                                                                                                                       & BN-CVAE                                                                                                                                    \\
1. okay bye - bye             & 1. okay so we'll go ahead and start                                                                                                                              & 1. okay bye - bye                                                                                                                          \\
2. bye                        & \begin{tabular}[c]{@{}l@{}}2. so i guess it depends on how much \\ you are you \end{tabular}                                                                     & 2. nice talking to you too                                                                                                                 \\
3. okay bye - bye             & 3. it's                                                                                                                                                          & 3. all right take care bye - bye                                                                                                           \\ 
\hline
\multicolumn{3}{l}{Topic: RECYCLING}                                                                                                                                                                                                                                                                                                          \\ 
\hline
\multicolumn{3}{l}{Context: are they doing a lot of recycling out in georgia}                                                                                                                                                                                                                                                                 \\ 
\hline
\multicolumn{3}{l}{\begin{tabular}[c]{@{}l@{}}Target (statement-non-opinion) : well at my workplace they are we have places for \\ aluminum cans and we have everybody's been unk a separate trash can for \end{tabular}}                                                                                                                     \\ 
\hline
CVAE                          & CVAE (BOW)                                                                                                                                                       & BN-CVAE                                                                                                                                    \\
1. yeah                       & \begin{tabular}[c]{@{}l@{}}1. well that's good for a while i'll tell you\\ that you're not doing anything at ti and \\ then you're in a small town \end{tabular} & \begin{tabular}[c]{@{}l@{}}1. well we do recycle \\ newspapers \end{tabular}                                                               \\
2. yeah                       & \begin{tabular}[c]{@{}l@{}}2. oh i know i've got a lot of trees and \\ trees and stuff and \end{tabular}                                                         & 2. yes i do too                                                                                                                            \\
3. yeah                       & \begin{tabular}[c]{@{}l@{}}3. yeah it's like you know people that \\ want to be unk and they're not going\\ to bother you to make a mess \end{tabular}           & \begin{tabular}[c]{@{}l@{}}3. well we're at a point where \\ we're going to be a landfill\\ space \end{tabular}                            \\ 
\hline
\multicolumn{3}{l}{Topic: UNIVERSAL HEALTH INS}                                                                                                                                                                                                                                                                                               \\ 
\hline
\multicolumn{3}{l}{Context: some of the good obviously that nobody has to worry about health care}                                                                                                                                                                                                                                            \\ 
\hline
\multicolumn{3}{l}{\begin{tabular}[c]{@{}l@{}}Target (statement-non-opinion) : and i guess i'll have to help with grandchildren\\ one of these days i hope \end{tabular}}                                                                                                                                                                     \\ 
\hline
CVAE                          & CVAE (BOW)                                                                                                                                                       & BN-CVAE                                                                                                                                    \\
1. um - hum                   & \begin{tabular}[c]{@{}l@{}}1. okay well see we don't have any \\ choice of any of those \end{tabular}                                                            & \begin{tabular}[c]{@{}l@{}}1. well i hope that we should \\ have a balanced budget \end{tabular}                                           \\
2. uh - huh                   & 2.um - hum                                                                                                                                                       & 2. uh - huh                                                                                                                                \\
3. uh - huh                   & 3. yeah                                                                                                                                                          & 3. well that's a good idea                                                                                                                 \\
\hline
\end{tabular}
\end{table*}

\end{document}